\documentclass{article}
\usepackage{ijcai19}

\usepackage{times}
\usepackage{soul}
\usepackage{url}
\usepackage[hidelinks]{hyperref}
\usepackage[utf8]{inputenc}
\usepackage[small]{caption}
\usepackage{graphicx}
\usepackage{amsmath}
\usepackage{amssymb}
\usepackage{amsthm}
\usepackage{booktabs}
\usepackage{algorithm}
\usepackage{algorithmic}
\usepackage{thm-restate}
\usepackage[utf8]{inputenc}

\newtheorem{definition}{Definition}

\title{Explanation Beyond Intuition: A Testable Criterion for Inherent Explainability}

\author{
Mike Merry
\and
Pat Riddle\And
Jim Warren
\affiliations
School of Computer Science, University of Auckland
\emails
\textit{m.merry@auckland.ac.nz},
\{pat,jim\}@cs.auckland.ac.nz
}

\begin{document}

\maketitle

\begin{abstract}

Inherent explainability is the gold standard in Explainable Artificial Intelligence (XAI). However, there is not a consistent definition or test to demonstrate inherent explainability. Work to date either characterises explainability through metrics, or appeals to intuition - "we know it when we see it".

We propose a globally applicable criterion for inherent explainability. The criterion uses graph theory for representing and decomposing models for \textit{structure-local explanation}, and recomposing them into global explanations. We form the structure-local explanations as \textit{annotations}, a verifiable hypothesis-evidence structure that allows for a range of explanatory methods to be used.

This criterion matches existing intuitions on inherent explainability, and provides justifications why a large regression model may not be explainable but a sparse neural network could be. We differentiate \textit{explainable} - a model that allows for explanation - and \textit{explained} - one that has a verified explanation. Finally, we provide a full explanation of PREDICT - a Cox proportional hazards model of cardiovascular disease risk, which is in active clinical use in New Zealand. It follows that PREDICT is inherently explainable.

This work provides structure to formalise other work on explainability, and allows regulators a flexible but rigorous test that can be used in compliance frameworks.

\end{abstract}

\section{Introduction}

The value of Explainable AI (XAI) in sensitive fields remains high, but there are still evident challenges defining explanation, and formal tests for explainability. For inherent explainability, where whole models—not just localised predictions—are fully explainable, the test remains predominantly intuitive - "we know it when we see it" \cite{doshi2017towards}.

This creates challenges for regulators and practitioners seeking safety and transparency in health, law, and insurance. Without an objective standard, any framework or regulation requiring explainability has a weak foundation. "We know it when we see it" is not auditable.

Here, we describe the mechanics of formalised explainability - functional graph decomposition of models, hypothesis-evidence structures called annotations, and composition of subgraphs with annotations. These mechanics are then used to formalise the test for inherent explainability. We demonstrate that this test aligns with existing claims and intuitions, and we demonstrate its application on a cardiovascular disease risk prediction model in use in clinical care.

\section{Background}

There is little consensus on what explainability in machine learning is and how to evaluate it for benchmarking \cite{doshi2017towards}. It is referred to ambiguously as interpretability, comprehensibility, understandability and explainability \cite{guidotti2018survey}. Although there is substantial work characterising and describing ways in which we approach it, even formalising them in taxonomies and similar, there is no definitive standard for inherent explainability \cite{doshi2017towards,adadi2018peeking,guidotti2018local,guidotti2018survey,ribeiro2018anchors,holzinger2017medical,samek2017explainable,lacave2002review}. 

The foundation of many techniques rely on a claim of inherent explainability, where some model classes, especially regression models and decision trees, are considered explainable axiomatically \cite{doshi2017towards}. Proxy techniques including LIME \cite{ribeiro2016why} and LORE \cite{guidotti2018local} create explanations of local predictions by reducing the complex model to a proxy model using inherently-explainable models. This class of approach naturally relies on the acceptance of the inherent explainability of the proxy models.

Other approaches include attention \cite{vaswani2017attention} or attribution based methods \cite{sundararajan2017axiomatic}. These methods provide the relative importance or criticality of differing features - the what of predictions - but they do not address specifically how the methods use these features to produce the result. This can help address some risks, such as identifying models that have been inadvertantly trained on incorrect features \cite{degrave2021ai}, so help reject classes of poor models, but they do not demonstrate how a model is behaving.

Much of the motivation for XAI is related to practical uses in sensitive contexts in real-world applications such as healthcare. Human-centric metrics like comprehension and trust have become increasingly important \cite{rago2018argumentation,merry2021mental}, as have user studies.

The importance of structures within (especially neural-network type) models, is increasing in prominence in XAI. Rago et al identify substructures within neural networks to explain through argumentation \cite{rago2018argumentation}. Olca et al. have explained transformer behaviour regarding substructures called "circuits" with methods such as ablation or manipulation of the model to determine the importance of the structure \cite{olah2020zoom}. GNNExplainer \cite{ying2019gnnexplainer} and similar work uses explanatory subgraphs to create instance-local explanations. These approaches share an intuition that model structure is explanatorily relevant, but lack a unifying framework for when structural decomposition constitutes a complete explanation

The work of miller \cite{miller2023evaluative} and Le et al \cite{le2024evaluative} have advocated for hypothesis-driven approaches to XAI, where systems provide evidence for and against hypotheses rather than direct recommendations. 

Despite this work, no formal criterion exists for testing inherent explainability claims—no standard that creators can demonstrate and regulators verify.

\section{A Formal Framework for Explanation}

Here, we build a formalisation of the mechanics needed for testing explainability. We start from a formal definition of explainable that addresses what needs to be demonstrated. We then build the mechanics of explanation in our framework: the atomic syntax of explanations - a hypothesis-evidence structure called annotation; the form in which they are applied to computational models - a computational graph formalisation from automatic differentiation; the construction method to create global explanations from structure-local explanations - composition of subgraphs; and finally the semantics of explanation - the methods by which evidence can be verified.

\subsection{Explainable}

\begin{definition}[Explanation]
\label{def:explanation}
An explanation of a model $M$ is a representation of $M$ in a language $L$, for an audience $A$, and for a purpose $P$. The triple $(A, L, P)$ constitutes the \emph{context} of the explanation.
\end{definition}

\begin{definition}[Explainable]
\label{def:explainable}
A model is explainable if there exists an explanation where the representation is accurate to the underlying model, the language is understandable by the audience, and the purpose of the explanation is met
\end{definition}

Merry et al. established that explanation must be understood relative to context (purpose, audience, language), arguing that existing definitions failed by treating explanation as context-independent \cite{merry2021mental}.

\subsection{Verifiable Hypotheses}

\begin{definition}[Verifiability]
\label{def:verifiability}
A claim about a model is verifiable if there exists a method, accepted by the intended audience as epistemically adequate, to determine whether the claim accurately characterises the model's behaviour.
\end{definition}

A central requirement of our framework is that explanatory claims be \emph{verifiable} - that there exist methods by which claims can be assessed for accuracy. We deliberately adopt a pluralistic stance on verification. Different  communities accept different forms of evidence and different standards of proof. This extends Definition~\ref{def:explainable}'s position that explanations are context dependent. This definition encompasses:
\begin{itemize}
    \item Mathematical derivation (proof that a claim follows from axioms)
    \item Empirical testing (experimental observation confirming predictions)
    \item Logical inference (deduction from established premises)
    \item Computational verification (algorithmic confirmation of properties)
\end{itemize}

The key constraint is that verification methods must be epistemically acceptable to the intended audience, just as the language must be understandable. This mirrors scientific peer review: different communities apply different standards, yet the process remains sound because standards are established and consistently applied.

\begin{definition}[Hypothesis-Evidence Structure]
\label{def:hypothesis-evidence}
An explanation consists of:
\begin{itemize}
    \item A \textbf{hypothesis}: A claim about what a model component does
    \item \textbf{Evidence}: Verification that the hypothesis accurately characterises the component's behaviour
\end{itemize}
\end{definition}

Explanations in our framework have a two-level structure that separates \textit{what} is claimed from \textit{how} it is verified. The hypothesis and evidence may be at different levels of technical sophistication. This separation serves a crucial function: it allows the same underlying explanation to serve different audiences. The hypothesis can be stated in accessible terms while the evidence provides rigorous grounding for those who require it.

Consider an analogy from medicine:

\begin{itemize}
    \item \textbf{Hypothesis}: ``Beta-blockers reduce performance anxiety by suppressing elevated heart rate.''
    \item \textbf{Evidence}: Pharmacokinetic studies, clinical trials, physiological measurements demonstrating the mechanism.
\end{itemize}

A patient can understand the hypothesis without understanding the evidence. A pharmacologist can verify the evidence without simplifying it. The explanation functions across audiences because hypothesis and evidence are separable.

Similarly, in model explanation:

\begin{itemize}
    \item \textbf{Hypothesis}: ``This subgraph creates a U-shaped risk curve over the age variable.''
    \item \textbf{Evidence}: Closed-form derivation showing the subgraph implements $f(x) = ax^2 + bx + c$; ablation study showing removal eliminates the non-linearity.
\end{itemize}

This creates the form of an explanation (hypothesis + explanation), and a test (verifiability), which aligns with the definition of explainability. We can now say that a model, or part of a model, is explainable if we can provide a verifiable hypothesis for its behaviour. We now present the structures to formalise "a model, or part of a model".

\subsection{Functions as Graphs: A Unified Representation}

We want a method to break larger models down into something smaller that we can explain, and combine together to make each explanatory step possible. We will use graph theory as the basis for this method, using subgraphs as the smaller structure to apply our annotations to.

Following the computational graph formalism standard in automatic differentiation \cite{griewank2008evaluating}, we represent models as directed acyclic graphs where nodes are operations and edges represent data flow. This representation applies uniformly to regression models, decision trees, and neural networks.

\begin{definition}[Computational Graph]
\label{def:computational-graph}
A computational graph $G = (V, E, W, \Phi)$ is a directed acyclic graph where:
\begin{itemize}
    \item $V = V_I \cup V_C \cup V_O$ partitions nodes into inputs (sources), computations (internal), and outputs (sinks)
    \item $E \subseteq V \times V$ represents directed data flow
    \item $W: E \rightarrow \mathbb{R}$ assigns edge weights (where applicable)
    \item $\Phi: V_C \cup V_O \rightarrow \mathcal{F}$ assigns an arbitrary function to each non-input node
\end{itemize}
\end{definition}

\noindent Given input values, the graph is evaluated by topological traversal:
\begin{equation}
a_v = \phi_v\left(\{(a_u, w_{uv}) : (u, v) \in E\}\right)
\end{equation}

\paragraph{Generality of $\Phi$.} Unlike standard neural network formalisms that restrict activation functions to a fixed set (ReLU, sigmoid, etc.), we permit \emph{any} function as a node operation. This generality enables unified treatment: in linear regression, $\Phi$ computes weighted sums; in decision trees, $\Phi$ evaluates threshold predicates; in neural networks, $\Phi$ applies activation functions to weighted sums. The framework thus captures all standard ML model classes within a single representation.

\paragraph{Examples.} A linear regression $y = \beta_0 + \sum_i \beta_i x_i$ is a graph with direct edges from each input $x_i$ to output $y$, edge weights $\beta_i$, and $\Phi(y)$ computing the weighted sum. A decision tree maps directly: internal nodes apply threshold predicates, edges represent branches, and leaves are outputs. A neural network has neurons as computation nodes, connection weights on edges, and $\Phi$ applying activation functions to weighted inputs.

\subsection{Formalised Explanations}

\begin{definition}[Annotation]
\label{def:annotation}
An \textbf{annotation} $A$ is a pair:
\[
A = (S, H_A, \Xi_A)
\]
where:
\begin{itemize}
    \item $S = (V_{\text{entry}}, V_{\text{exit}}, V_S, E_S)$ is a valid subgraph
    \item $H_A$ is the \textbf{hypothesis}: a human-comprehensible claim about the subgraph's function
    \item $\Xi_A$ is the \textbf{evidence}: verification supporting $H_A$
\end{itemize}
\end{definition}

An \emph{annotation} extends a subgraph with meaning. The subgraph identifies \emph{the part of the model} that is being explained; the hypothesis states \emph{what it does}; the evidence demonstrates \emph{that this is true}.

The hypothesis-evidence structure is not an organisational layer on top of existing explanatory methods. Many artefacts currently termed explanations (e.g., feature attribution and attention) would fail to constitute explanations as they do not, by themselves, form a hypothesis about model behaviour. They may act as points of evidence to support a hypothesis, but are not complete by themselves, requiring additional work to meet this standard.

It is important to note that the evidence required here is only related to whether or not the hypothesis is a true claim about the model's behaviour, and is not about the validity of the model to the world. It is entirely possible to create a full and valid explanation of a model that is entirely specious with regards to the real world. A model can be inherently explainable but wrong.

\subsubsection{Compositional Explanation Structure}

Structural coverage alone is insufficient for a complete explanation. Knowing what each subgraph does individually does not tell us how they combine. \textbf{Composition itself requires explanation.}

Consider a model where:
\begin{itemize}
    \item Annotation $A_1$ explains: ``Subgraph $S_1$ computes age risk factor $f(\text{age})$''
    \item Annotation $A_2$ explains: ``Subgraph $S_2$ computes BP (blood pressure) risk factor $g(\text{BP})$''
\end{itemize}

Even with full structural coverage, we don't know:
\begin{itemize}
    \item Is the output $f(\text{age}) + g(\text{BP})$? (additive)
    \item Is it $f(\text{age}) \times g(\text{BP})$? (multiplicative)
    \item Is there an interaction term?
\end{itemize}

The combination requires its own explanation.

\subsubsection{Leaf and Composition Annotations}

We distinguish two types of annotations:

\begin{definition}[Leaf Annotation]
A \textbf{leaf annotation} is an annotation as previously defined (Definition~\ref{def:annotation}) - it explains a primitive subgraph of the model in isolation.
\end{definition}

\begin{definition}[Composition Annotation]
Given annotations $A_1, A_2, \ldots, A_k$ (leaf or composition), a \textbf{composition annotation} $C(A_1, \ldots, A_k)$ is a tuple:
\[
C = (S_C, H_C, \Xi_C, \{A_1, \ldots, A_k\})
\]
where:
\begin{itemize}
    \item $S_C = (V_{\text{entry}}^C, V_{\text{exit}}^C, V_C, E_C)$ is the \textbf{junction subgraph} where component outputs meet
    \item $H_C$ is the \textbf{composition hypothesis}: how component behaviours combine
    \item $\Xi_C$ is the evidence supporting $H_C$
    \item $\{A_1, \ldots, A_k\}$ are the \textbf{child annotations} being composed
\end{itemize}
\end{definition}

Composition annotations follow the same hypothesis-evidence structure and verification methods as leaf annotations.

\subsubsection{Annotation Hierarchy}

\begin{definition}[Annotation Hierarchy]
An \textbf{annotation hierarchy} $\mathcal{H}$ for a model $G$ is a rooted tree where:
\begin{itemize}
    \item Leaves are leaf annotations (explain primitive subgraphs)
    \item Internal nodes are composition annotations (explain how children combine)
    \item The root covers the global model (entry = inputs, exit = outputs)
\end{itemize}
\end{definition}

The hierarchy represents a \textbf{composition path} - the order in which local explanations combine into a global explanation. The order in which annotations are composed is important. If you have already explained that the combination of two effects creates a U-shaped response function, then a later explanation can use the U-shaped response without considering the component parts.

\subsection{Two Modes of Verification}

While our framework accepts any epistemically adequate verification method, two forms are particularly well-established for machine learning models.

\subsubsection{Analytical Verification}

Analytical verification establishes claims through mathematical derivation. Given a subgraph, we derive a closed-form expression for its input-output behaviour and verify that this expression has the properties claimed in the hypothesis.

\paragraph{Form:} ``The behaviour of subgraph $S$ is characterised by mathematical function $f$.''

\paragraph{Method:}
\begin{enumerate}
    \item Derive the closed-form expression for the subgraph by symbolic computation through the graph
    \item Analyse the mathematical properties of the resulting expression (linearity, monotonicity, inflection points, domain-range relationships)
    \item Verify that these properties match the hypothesis
\end{enumerate}

When tractable, analytical verification provides the strongest form of evidence. A mathematical derivation does not merely show that a claim \emph{happens to be true} for observed cases - it shows \emph{why} the claim must be true given the structure of the subgraph. Save that, in practice, mathematicians make mistakes, it provides the deepest explanation of models. This explanatory depth is particularly valuable when the audience includes those who can follow mathematical reasoning, but it may not be appropriate for most audiences.

\paragraph{Limitations:} Analytical verification is tractable only for sufficiently simple or small subgraphs. As subgraph complexity grows (more nodes, non-linear activation functions, complex compositions), closed-form expressions become unwieldy or impossible to derive. In such cases, we turn to empirical methods.

\paragraph{Example:} For a subgraph implementing $f(\text{age}) = \beta_1(\text{age} - 50)^2 + \beta_0$:
\begin{itemize}
    \item \emph{Hypothesis}: ``This subgraph creates a U-shaped risk curve with minimum at age 50''
    \item \emph{Evidence}: The derivative $f'(\text{age}) = 2\beta_1(\text{age} - 50)$ equals zero at age 50; $f''(\text{age}) = 2\beta_1 > 0$ confirms a minimum
\end{itemize}

\subsubsection{Empirical Verification}

Empirical verification establishes claims through experimental observation. We intervene on the model - removing, modifying, or isolating the subgraph - and observe how behaviour changes.

\paragraph{Form:} ``Subgraph $S$ contributes function $F$ to the model's behaviour.''

\paragraph{Method:} Design and execute experiments that test the hypothesis about subgraph behaviour. Common approaches include:

\begin{itemize}
    \item \textbf{Ablation}: Remove or zero-out the subgraph to confirm contribution to overall behaviour
    \item \textbf{Perturbation}: Systematically vary subgraph parameters to characterise sensitivity
    \item \textbf{Input-output sampling}: Sample inputs, record outputs, fit interpretable model to approximate functional form
    \item \textbf{Substitution}: Replace subgraph with interpretable equivalent to validate characterisation
\end{itemize}

Empirical methods extend the reach of verification beyond what analytical methods can handle. For complex subgraphs where closed-form derivation is intractable, experimental observation can still provide evidence for or against hypotheses about subgraph behaviour. Empirical verification also connects directly to observable model behaviour, grounding abstract claims in measurable outcomes.

\paragraph{Limitations:} Empirical verification is inherently statistical and observational. It shows that a claim is \emph{consistent with} observed behaviour, not that it \emph{must} hold. Edge cases or unusual inputs might reveal behaviours not captured by the hypothesis. However, for practical purposes, empirical evidence that covers the relevant input domain often provides sufficient grounds for accepting a hypothesis.

\paragraph{Relationship to instance-local methods:} The empirical techniques listed above - particularly input-output sampling and surrogate model fitting - are related to methods used in instance-local explanation (LIME, SHAP). The difference is the level of analysis: instance-local methods apply these techniques to explain individual predictions; we apply them to explain subgraph behaviour across all inputs. The same underlying methodology serves different explanatory purposes.
\subsection{Combined Coverage and Explainability}

\subsubsection{Structural and Compositional Coverage}

We distinguish two types of coverage required for a complete explanation:

\begin{definition}[Structural Coverage]
A set of leaf annotations $\mathcal{A}_{\text{leaf}}$ achieves \textbf{full structural coverage} of graph $G$ if every node in $G$ is covered:
\[
\forall v \in V: \text{covered}_{\mathcal{A}_{\text{leaf}}}(v)
\]
\end{definition}

\begin{definition}[Compositional Coverage]
An annotation hierarchy $\mathcal{H}$ achieves \textbf{full compositional coverage} if every required composition has a valid composition annotation explaining how its children combine.
\end{definition}

\noindent Structural coverage ensures all parts are explained. Compositional coverage ensures all combinations are explained. Neither alone is sufficient: structural coverage without compositional coverage means we have explained the parts but not how they interact; compositional coverage without structural coverage means we have explained combinations of things we haven't individually explained.

\subsubsection{Well-Formed Explanation}

\begin{definition}[Well-Formed Global Explanation]
\label{def:well-formed-global-explanation}
A global explanation $\mathcal{E} = (\mathcal{H}, \mathcal{A})$ is \textbf{well-formed} if:
\begin{enumerate}
    \item Every leaf annotation in $\mathcal{H}$ is valid
    \item Full structural coverage is achieved
    \item Full compositional coverage is achieved
    \item The root annotation covers the global model (entry = model inputs, exit = model outputs)
\end{enumerate}
\end{definition}

\noindent A well-formed explanation provides a complete account of the model: every component is explained, every composition is explained, and the hierarchy terminates in a global explanation.

A model is fully explained only when \textbf{both} structural and compositional coverage are complete.

\subsubsection{Explainability vs.\ Explainedness}

We distinguish two related but distinct concepts:

\begin{definition}[Inherent Explainability]
A model $G$ is \textbf{inherently explainable} iff there exists a well-formed global explanation $\mathcal{E}$ for $G$.
\end{definition}

\begin{definition}[Explainedness]
The \textbf{explainedness} of a model $G$ with respect to hierarchy $\mathcal{H}$ is the pair $(C_V^{\text{struct}}(\mathcal{H}), C_V^{\text{comp}}(\mathcal{H}))$.
\end{definition}

A model may be inherently explainable but not yet explained (work remains). We can use coverage to describe what work has been done, and what work remains.

\section{Inherent Explainability}

\subsection{The Test for Explainability}

We propose a sufficient condition for inherent explainability based on graph decomposition with compositional annotation. This is explicitly \textit{not a necessary condition} - we do not claim that models failing this criterion are unexplainable, only that models meeting it demonstrably are.

\begin{restatable}{criterion}{explainability}
A model $G$ is \textbf{inherently explainable} if there exists a well-formed global explanation $\mathcal{E} = (\mathcal{H}, \mathcal{A})$ such that:
\begin{enumerate}
    \item Every leaf annotation in $\mathcal{H}$ is valid
    \item Full structural coverage is achieved: every node in $G$ is covered by a leaf annotation
    \item Full compositional coverage is achieved: every required composition has a valid composition annotation
    \item The root annotation covers the global model (entry = model inputs, exit = model outputs)
\end{enumerate}
\end{restatable}

Both coverage types are required. Structural coverage without compositional coverage means we have explained all the parts but not how they combine. Compositional coverage without structural coverage means we have explained combinations of things we haven't individually explained.

\subsection{Validation: Linear Regression}

Consider linear regression: $y = \beta_0 + \beta_1 x_1 + \beta_2 x_2 + \ldots + \beta_n x_n$

\paragraph{Graph structure:} Each input $x_i$ connects directly to output $y$. Each edge carries weight $\beta_i$.

\paragraph{Leaf annotations:} For each feature, create annotation $A_i$ with:
\begin{itemize}
    \item Entry: $\{x_i\}$, Exit: $\{y\}$
    \item Hypothesis: ``Feature $x_i$ contributes $\beta_i$ per unit change''
    \item Evidence: Coefficient value from model
\end{itemize}

\paragraph{Structural coverage:} Each input node has exactly one outgoing edge, contained in its annotation. All nodes covered. $C_V^{\text{struct}} = 1$.

\paragraph{Composition:} How do the $n$ feature contributions combine? The composition annotation states: ``All features combine additively: $y = \sum_i \beta_i x_i + \beta_0$'', and we see that these features are important due to these coefficients.

For small $n$, this composition is comprehensible. A 5-variable regression can be understood as ``these five factors add up.'' The framework validates what intuition suggests: simple linear regression is explainable.

\subsection{The Large Regression Problem: When Compositional Coverage Fails}

Now consider a regression with 10,000 features: $y = \sum_{i=1}^{10000} \beta_i x_i + \beta_0$ Each feature gets a leaf annotation: ``$x_i$ contributes $\beta_i$ per unit.'' With 10,000 such annotations, we achieve coverage of all nodes.

To construct a well-formed explanation, we need a composition annotation explaining how 10,000 leaf explanations combine into a global understanding. What would such a composition look like?

\paragraph{Option 1: Single 10,000-ary composition.} We could create one composition annotation: ``All 10,000 features combine additively.'' This is structurally valid - it correctly describes how the model computes its output. But does it constitute an \emph{explanation} in the sense of Definition~\ref{def:explainable}?

Recall that an explanation must be ``understandable by an audience for a purpose.'' A claim that 10,000 features combine additively, while true, provides no insight beyond restating that the model is linear. It does not help an audience understand \emph{why} the model makes particular predictions, identify which features matter, or gain any actionable understanding. For most audiences and purposes, this ``explanation'' fails to explain.

The composition annotation is formally valid but explanatorily vacuous. It is well-formed by Definition~\ref{def:well-formed-global-explanation}, but would most-likely fail the audience test under Definition~\ref{def:explainable}. It is explainable (theoretically), but until it is explained in a way that meets Definition~\ref{def:explainable}, this is not useful.

\paragraph{Option 2: Hierarchical composition through meaningful groupings.} A more promising approach would group features into interpretable clusters - demographics, lab values, medications, lifestyle factors - then compose within groups before composing across groups. For example:
\begin{itemize}
    \item First level: ``Age, sex, and ethnicity combine to form demographic risk''
    \item Second level: ``BP, cholesterol, and glucose combine to form cardiovascular indicators''
    \item Third level: ``Demographic risk and cardiovascular indicators combine additively with interaction''
\end{itemize}

This hierarchical structure could yield a comprehensible global explanation. But it requires that meaningful groupings \emph{exist}. For 10,000 arbitrary features - perhaps generated by automated feature engineering or extracted from high-dimensional data - such groupings may not exist. The features may have no natural clustering that would support interpretable composition.

When meaningful groupings exist, the model may be explainable despite having many features. When they do not, compositional coverage fails even though structural coverage is trivial.

\paragraph{Resolution:} The framework resolves what might otherwise be a paradox. Regression is commonly claimed to be ``inherently explainable,'' yet a 10,000-coefficient regression clearly is most likely not explainable in any useful sense (there may be exceptions). Our framework shows why: structural coverage (explaining each coefficient) is easy, but compositional coverage (forming a comprehensible global picture) is likely impossible. The distinction between the two types of coverage captures what intuition grasps but previous formalisms could not express.

If an author wishes to claim explainability for a large regression model, there is a clear criterion: demonstrate a valid composition path. Should the author be able to do this, and the evidence is considered valid, then the model can be considered not just explainable, but explained.

\subsection{Validation: Decision Trees}

Decision trees are widely considered inherently explainable. Our framework validates this:

\paragraph{Graph structure:} Internal nodes are decision predicates; leaf nodes are predictions. Each root-to-leaf path is a conjunction of conditions.

\paragraph{Leaf annotations:} Each path gets an annotation:
\begin{itemize}
    \item Entry: root, Exit: leaf
    \item Hypothesis: ``If $C_1 \land C_2 \land \ldots \land C_k$ then predict $y$''
    \item Evidence: The path structure itself
\end{itemize}

\paragraph{Structural coverage:} Every node lies on at least one path. All paths annotated means all nodes covered.

\paragraph{Composition:} Paths are mutually exclusive (only one fires for any input). The composition annotation states: ``Exactly one path matches any input; the prediction is that path's leaf value.'' This is a simple, comprehensible composition.

It may be that we find creating the hypothesis for a given path the more challenging task depending on the model. If, when explaining a model, one is not able to form the hypothesis then there is no explanation and it fails the test.

\subsection{Why Dense Networks Fail}

Dense neural networks are commonly described as ``black boxes.'' Our framework provides formal grounding for this intuition.

Consider a modest dense network: 4 inputs, 2 hidden layers of 16 neurons each, and 1 output. This network has:
\begin{itemize}
    \item 37 nodes (4 input + 16 + 16 + 1 output)
    \item 336 edges (4$\times$16 + 16$\times$16 + 16$\times$1)
\end{itemize}

To decompose this network into meaningful subgraphs, we would need to identify substructures where behaviour can be characterised independently. But in a dense network, every neuron in layer $l$ connects to every neuron in layer $l+1$. There are no natural boundaries.

\textbf{The identification problem:} To construct a well-formed 
explanation, we must identify subgraphs that have meaningful, 
independent functions---subgraphs for which a coherent hypothesis 
can be formed. Not every subgraph is meaningful; most arbitrary 
subsets of a network's edges do not compute anything interpretable 
in isolation.

The space of candidate subgraphs is $2^{|E|}$---all possible subsets 
of edges. For our 4-16-16-1 network with 336 edges, this is 
$2^{336} \approx 1.4 \times 10^{101}$, more than a googol. Even if 
meaningful subgraphs exist somewhere in this space, finding them is 
intractable.

\textbf{Why this differs from sparse networks:} In a sparse network, 
selective connectivity creates natural boundaries that suggest where 
meaningful subgraphs reside. A neuron connecting to only 3 of 16 
neurons in the next layer architecturally delineates a substructure. 
The decomposition is guided by the structure itself.

In a dense network, every neuron connects to every other neuron. 
There are no architectural hints about where meaning might reside. 
One must search for needles in a $10^{101}$-sized haystack---assuming 
the needles exist at all.

\section{Case Study Summary: PREDICT Cardiovascular Risk Model}
\begin{figure*}[t]
    \centering
    \includegraphics[width=\textwidth]{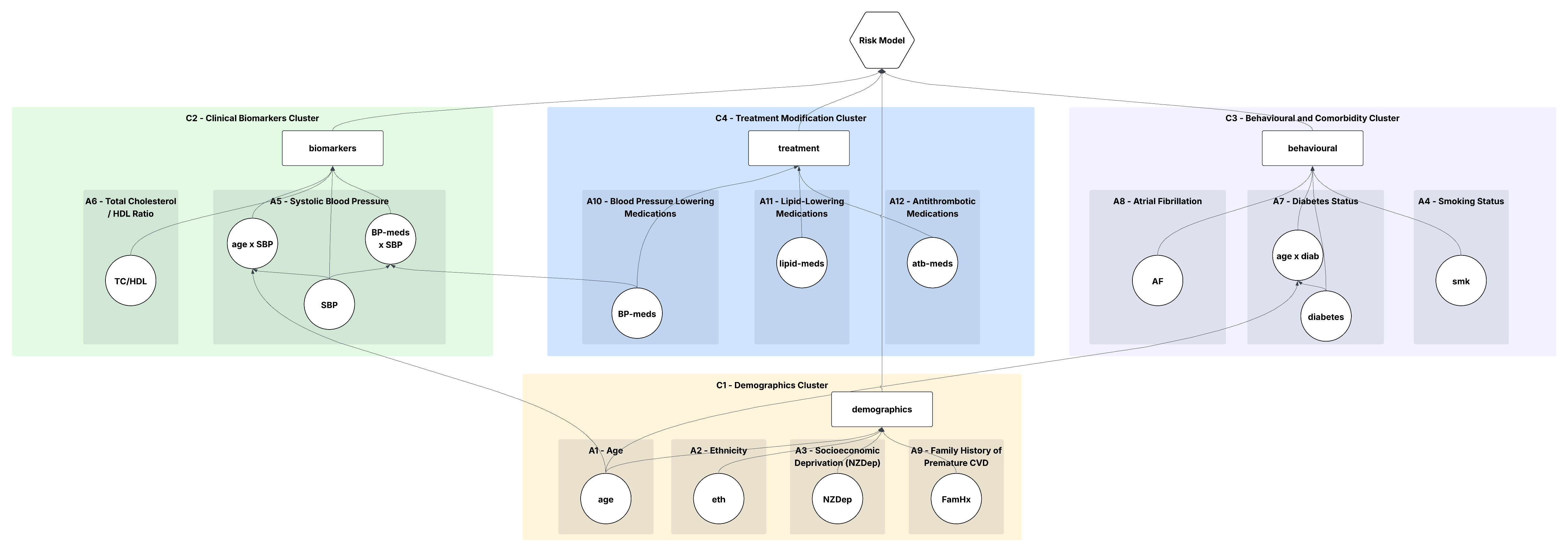}
    \caption{Overview of the PREDICT cardiovascular risk model explanation structure, showing the hierarchical organization of leaf annotations, composition annotations, and global composition.}
    \label{fig:predict-overview}
\end{figure*}
To demonstrate the framework's applicability, we present in Appendix A a complete explanation of the PREDICT cardiovascular disease (CVD) risk model - a Cox proportional hazards model currently deployed in approximately one-third of New Zealand primary care practices \cite{pylypchuk2018cvd,kerr2019}. The model predicts 5-year CVD risk using 12 predictor variables and 3 interaction terms.

We constructed a well-formed global explanation consisting of:
\begin{itemize}
    \item \textbf{12 leaf annotations} covering each predictor (age, ethnicity, deprivation, smoking status, blood pressure, cholesterol ratio, diabetes, atrial fibrillation, family history, and three medication categories)
    \item \textbf{4 composition annotations} grouping predictors into clinically meaningful clusters (demographics, biomarkers, behavioral/comorbidities, treatment effects)
    \item \textbf{1 global composition} explaining the additive combination on the log-hazard scale and transformation to absolute risk
\end{itemize}

Each annotation follows the hypothesis-evidence structure. For example, the age annotation hypothesises that each additional year increases CVD risk by approximately 7--8\% (HR $\approx$ 1.07--1.08), with evidence from analytical derivation of the coefficient and its exponential. We intend the annotations to be for academics in computational areas with experience with predictive models. The complete explanation achieves full structural and compositional coverage, satisfying the explainability criterion.

The explanation was initially drafted using generative AI tools and subsequently verified manually by the authors against the original publications. We consider this workflow valid - the framework places no constraints on how explanations are produced, only on their verifiability. It also serves as an example of how we might construct tools to support this framework.

\paragraph{Readers as verifiers.} We observe that readers of this paper are themselves acting to verify the explanation as the framework describes. In assessing whether the PREDICT explanation is accurate, comprehensible, and complete, readers perform audience-relative verification using methods they accept as epistemically adequate (mathematical derivation, consistency with published sources). This is not circular; rather, it demonstrates that the framework is compatible with established scientific practice. Acceptance of the case study as valid constitutes empirical evidence that the framework can produce verifiable explanations for models of meaningful complexity.

\section{Discussion}

This paper provides several contributions to the theory and practice of XAI. Most importantly, it provides a rigorous and testable definition for inherent explainability. Where previous work has relied on informal but broadly accepted claims, our framework provides the precise criteria.

We have shown that the distinction between structural and compositional coverage is particularly important. It provides the language to describe why the boundaries of inherent explainability exist where they do. Extremely large regression models are not inherently explainable because creating composition paths of leaf explanations is challenging. Dense neural networks are not inherently explainable because there is no tractable decomposition for explanation.

The corollary of our work is that proxy based methods which depend on the inherent explainability of their surrogate models are on a stronger footing. Rather than an appeal to the axiomatic inherent explainability of other models, it is possible to test that foundation. This framework also values attention- and attribution-based methods, as well as other post-hoc methods, as being the evidence base for hypotheses. For structural approaches, such as transformer circuits and GNNExplainer, we have provided a formalised framework that helps justify their methods, and can help to create consistency. It provides the additional steps of determining what is required for full versus partial explanation. 

We have used the term tractable loosely. This is because it is unclear what the upper bound on what is tractable. Although we have presented this method starting from the decomposition of models to leaf annotations, the explanatory method is by construction through composition. One needn't start from a large model, but could start from a single explainable leaf and progressively add complexity while retaining explainability. The effort lies not in decomposition but in constructing evidence for annotations. We do not claim to extend explainability bounds, but where claims are made—such as transformer circuits—this provides formal reference.

The case study shows the boundary reaches at least Cox proportional hazards; we expect it is substantially higher. As the onus of proof is on the person creating the model, rather than artificially constrain any idea of what is tractable, we allow it to simply be what an author can demonstrate that is accepted by the audience. Ultimately, the boundary of what is tractable is empircal - either you can present an epistemically valid explanation, or not.

This framework provides a scalable level of rigour for explainability - which is that standards for verifiability are context dependent but can be set in a similar way as it is in scientific communities. It can accept our informal intuitions and it can support regulatory frameworks - we can demand high levels of verifiable evidence for critical claims. It also supports multiple explanations for multiple audiences so that a statistician, a clinician and a patient can all have their appropriate explanations in the same way they can regarding the explanation of a drugs behaviour.

The framework also provides a higher order structure for considering other explainability metrics such as complexity, graph depth, and similar. Rather than being a test of explainability, other explainability metrics provide heuristics for when a model might be fully explained. As depth increases, compositional paths become harder to construct. This may also provide guidance for tractability. It may be that we learn that explanations for specific audiences have an upper bound on given metrics which can help us in creating tools for building explanations that are accepted and valid.

However, this framework does not specifically address how to evaluate the quality of an explanation and if it is useful. This is similarly context-dependent on both the audience and purpose. Programmatically-generated explanations may well be valid, but if a patient wants an explanation to guide clinical decision making, then the model creator must verify that the explanation is good in that context.

The use of computational graphs leaves open future research directions to explain neural networks. Convolutional, recurrent and transformer models are all representable in this fashion. The work on circuits in transformers \cite{olah2020zoom} uses similar structural elements of their models to drive explanation. Our own work on NEAT networks also demonstrates that evolution of sparse networks can produce performant models, and these are, by design, tractable for subgraph analysis. We believe that this could be an effective pathway for creating explainable neural networks.

However, this framework does not address the costs in developing explanations. Generating annotations, verifying them, and demonstrating that these are acceptable has a high time cost and there are no current tools to support or ease this. Although this creates a rigorous method, we expect that practical applications may simplify or allow for approximate implementations of the test, rather than a fully rigorous approach. That a fully rigorous approach is available is still valuable, even if used rarely, as any gap in rigour can then be evaluated by the audience.

Further, it may be the case that only a proportion of a model is explained. It would be reasonable to create metrics describing the amount of a model that is explained, for example by the proportion of the structure, or by the amount of variance in the resulting model that substructure is responsible for within a dataset. These may allow for descriptors of partial explainability where 90\% of the structure, or 100\% of the variance caused by a specific variable is explained. We would caution the use of such metrics; knowing which x\% is the most important to explain is difficult to know a-priori.

\section{Conclusion}

We have proposed a formal framework for compositional explanation of models. Using this framework, we have been able to define a sufficient criterion for inherent explainability that matches existing intuitions. The framework:

\begin{enumerate}
    \item \textbf{Operationalises contextual explanation} through a methodology of decomposition, local explanation, and composition
    \item \textbf{Distinguishes structure-local explanation} from instance-local explanation
    \item \textbf{Formalises coverage} at two levels: structural and compositional
    \item \textbf{Introduces annotation hierarchies} distinguishing leaf annotations from composition annotations
\end{enumerate}

The key insights are that ''inherent explainability'' can be expressed as structural explanation, that claims of inherent explainability typically describe the explanation of atomic parts of the structure, and that composition of these parts requires explanation. When combined, we can express why simple regression is explainable while very large regression models are not, and why sparse neural networks may well be, but most dense neural networks aren't.

We believe that this work provides a better framework for testing explainability for use in both academia and in practice than has been previously used, and that the insights will lead to new avenues of developing explainable models.

\bibliographystyle{named}
\bibliography{ijcai19}

@article{doshi2017towards,
  title={Towards a rigorous science of interpretable machine learning},
  author={Doshi-Velez, Finale and Kim, Been},
  journal={arXiv preprint arXiv:1702.08608},
  year={2017}
}

@article{adadi2018peeking,
  title={Peeking inside the black-box: A survey on explainable artificial intelligence (XAI)},
  author={Adadi, Amina and Berrada, Mohammed},
  journal={IEEE Access},
  volume={6},
  pages={52138--52160},
  year={2018},
  publisher={IEEE}
}

@article{guidotti2018survey,
  title={A survey of methods for explaining black box models},
  author={Guidotti, Riccardo and Monreale, Anna and Ruggieri, Salvatore and Turini, Franco and Giannotti, Fosca and Pedreschi, Dino},
  journal={ACM computing surveys (CSUR)},
  volume={51},
  number={5},
  pages={1--42},
  year={2018},
  publisher={ACM New York, NY, USA}
}

@inproceedings{ribeiro2018anchors,
  title={Anchors: High-precision model-agnostic explanations},
  author={Ribeiro, Marco Tulio and Singh, Sameer and Guestrin, Carlos},
  booktitle={Proceedings of the AAAI conference on artificial intelligence},
  volume={32},
  year={2018}
}

@article{holzinger2017medical,
  title={What do we need to build explainable AI systems for the medical domain?},
  author={Holzinger, Andreas and Biemann, Chris and Pattichis, Constantinos S and Kell, Douglas B},
  journal={arXiv preprint arXiv:1712.09923},
  year={2017}
}

@article{samek2017explainable,
  title={Explainable artificial intelligence: Understanding, visualizing and interpreting deep learning models},
  author={Samek, Wojciech and Wiegand, Thomas and M{\"u}ller, Klaus-Robert},
  journal={arXiv preprint arXiv:1708.08296},
  year={2017}
}

@article{lacave2002review,
  title={A review of explanation methods for Bayesian networks},
  author={Lacave, Carmen and D{\'\i}ez, Francisco J},
  journal={The Knowledge Engineering Review},
  volume={17},
  number={2},
  pages={107--127},
  year={2002},
  publisher={Cambridge University Press}
}

@inproceedings{vaswani2017attention,
  title={Attention is all you need},
  author={Vaswani, Ashish and Shazeer, Noam and Parmar, Niki and Uszkoreit, Jakob and Jones, Llion and Gomez, Aidan N and Kaiser, {\L}ukasz and Polosukhin, Illia},
  booktitle={Advances in neural information processing systems},
  volume={30},
  year={2017}
}

@inproceedings{sundararajan2017axiomatic,
  title={Axiomatic attribution for deep networks},
  author={Sundararajan, Mukund and Taly, Ankur and Yan, Qiqi},
  booktitle={International conference on machine learning},
  pages={3319--3328},
  year={2017},
  organization={PMLR}
}

@inproceedings{rago2018argumentation,
  title={Argumentation-based recommendations: Fantastic explanations and how to find them},
  author={Rago, Antonio and Cocarascu, Oana and Bechlivanidis, Christos and Toni, Francesca},
  booktitle={Proceedings of the 27th International Joint Conference on Artificial Intelligence},
  pages={5049--5055},
  year={2018}
}

@article{olah2020zoom,
  title={Zoom in: An introduction to circuits},
  author={Olah, Chris and Cammarata, Nick and Schubert, Ludwig and Goh, Gabriel and Petrov, Michael and Carter, Shan},
  journal={Distill},
  volume={5},
  number={3},
  pages={e00024--001},
  year={2020}
}

@article{degrave2021ai,
  title        = {AI for radiographic COVID-19 detection selects shortcuts over signal},
  author       = {DeGrave, Alex J. and Janizek, Joseph D. and Lee, Su-In},
  journal      = {Nature Machine Intelligence},
  volume       = {3},
  number       = {7},
  pages        = {610--619},
  year         = {2021},
  publisher    = {Nature Publishing Group},
  doi          = {10.1038/s42256-021-00338-7}
}

@article{guidotti2018local,
  author    = {Guidotti, Riccardo and Monreale, Anna and Ruggieri, Salvatore and Turini, Franco and Giannotti, Fosca and Pedreschi, Dino},
  title     = {Local Rule-Based Explanations of Black Box Decision Systems},
  journal   = {arXiv preprint arXiv:1805.10820},
  year      = {2018}
}

@inproceedings{le2024evaluative,
  author    = {Le, Ty and Miller, Tim and Singh, Ronal and Sonenberg, Liz},
  title     = {Evaluative {AI}: An Empirical Evaluation of Evaluative {AI}},
  booktitle = {Proceedings of the 2024 CHI Conference on Human Factors in Computing Systems},
  year      = {2024}
}

@article{merry2021mental,
  author    = {Merry, Mike and Riddle, Pat and Warren, Jim},
  title     = {A Mental Models Approach for Defining Explainable Artificial Intelligence},
  journal   = {BMC Medical Informatics and Decision Making},
  volume    = {21},
  number    = {344},
  year      = {2021}
}

@article{miller2023evaluative,
  author    = {Miller, Tim},
  title     = {Evaluative {AI}},
  journal   = {arXiv preprint arXiv:2301.01785},
  year      = {2023}
}

@inproceedings{ribeiro2016why,
  author    = {Ribeiro, Marco Tulio and Singh, Sameer and Guestrin, Carlos},
  title     = {``{W}hy Should {I} Trust You?'': Explaining the Predictions of Any Classifier},
  booktitle = {Proceedings of the 22nd ACM SIGKDD International Conference on Knowledge Discovery and Data Mining},
  pages     = {1135--1144},
  year      = {2016}
}

@inproceedings{ying2019gnnexplainer,
  author    = {Ying, Zhitao and Bourgeois, Dylan and You, Jiaxuan and Zitnik, Marinka and Leskovec, Jure},
  title     = {{GNNE}xplainer: Generating Explanations for Graph Neural Networks},
  booktitle = {Advances in Neural Information Processing Systems},
  volume    = {32},
  year      = {2019}
}

@book{griewank2008evaluating,
  author    = {Griewank, Andreas and Walther, Andrea},
  title     = {Evaluating Derivatives: Principles and Techniques of Algorithmic Differentiation},
  publisher = {SIAM},
  edition   = {2nd},
  year      = {2008}
}

@article{pylypchuk2018cvd,
  author    = {Pylypchuk, Romana and Wells, Sue and Kerr, Andrew and others},
  title     = {Cardiovascular Disease Risk Prediction Equations in 400,000 Primary Care Patients in {N}ew {Z}ealand: A Derivation and Validation Study},
  journal   = {Lancet},
  volume    = {391},
  number    = {10133},
  pages     = {1897--1907},
  year      = {2018}
}

@article{kerr2019,
  author    = {Kerr, Andrew and Wells, Sue and Moffitt, Allan and Lund, Mayanna and Kreichbaum, Jim and Harwood, Matire and Jackson, Rod},
  title     = {A unified national cardiovascular disease ({CVD}) risk generator is required to address equity in the management of {CVD} risk in clinical practice in {New Zealand}},
  journal   = {New Zealand Medical Journal},
  year      = {2019},
  volume    = {132},
  number    = {1500},
  pages     = {89--94},
  pmid      = {31415503}
}

\appendix

\section{Case Study: PREDICT Cardiovascular Risk Assessment}
\label{sec:predict-case-study}

To demonstrate this approach, we present an explanation of an existing risk prediction model for cardivascular disease that is currently in use in New Zealand. We utilised LLM-based tools to accelerate the initial creation of annotations. We have manually refined the explanation, ensuring accuracy to the original papers and correct formation in the framework. We manually reviewed all claims and the validity of the explanation to the underlying model, ensuring that it meets the requirements of the criterion.

\subsection{Model Description}

The PREDICT cardiovascular disease risk assessment model calculates 5-year absolute risk of CVD events (fatal or non-fatal hospitalisations from ischemic heart disease, cerebrovascular events, peripheral vascular disease, or congestive heart failure) for New Zealanders aged 30--74 years without prior CVD, renal disease, or heart failure \cite{pylypchuk2018cvd}. The model was derived from 401,752 primary care patients with 15,386 CVD events during 1,685,521 person-years of follow-up. It is deployed in approximately one-third of New Zealand primary care practices and has been used to assess over 500,000 individuals \cite{kerr2019}.

The model uses sex-specific Cox proportional hazards regression with 12 predictor variables plus 3 interaction terms, making it tractable yet realistic in complexity for clinical decision support.

\subsection{Model Structure}

\paragraph{Predictor variables:}
\begin{enumerate}
    \item Age (continuous, centered at mean: 56.1y women, 51.8y men)
    \item Ethnicity (categorical: European [ref], M\=aori, Pacific, Indian, Chinese/Asian)
    \item NZ Deprivation Index quintile (continuous, 1--5)
    \item Smoking status (categorical: never [ref], ex-smoker, current smoker)
    \item Family history of premature CVD (binary)
    \item Atrial fibrillation diagnosis (binary)
    \item Diabetes status (binary)
    \item Systolic blood pressure (continuous, centered at $\sim$129 mmHg)
    \item Total cholesterol/HDL ratio (continuous, centered at $\sim$3.7 women, $\sim$4.4 men)
    \item On blood pressure-lowering medication (binary)
    \item On lipid-lowering medication (binary)
    \item On antithrombotic medication (binary)
\end{enumerate}

\paragraph{Interaction terms:} 
\begin{enumerate}
    \item Age $\times$ Diabetes
    \item Age $\times$ Systolic BP
    \item BP-lowering medication $\times$ Systolic BP
\end{enumerate}

\paragraph{Full model}

For each sex (women/men), the 5-year absolute cardiovascular risk is given by
\begin{equation}
\text{Risk}_5(x)
=
1 - \bigl(S_0^{\text{sex}}(5)\bigr)^{\exp\bigl(\eta(x)\bigr)}
\end{equation}
where \(S_0^{\text{sex}}(5)\) is the sex-specific baseline 5-year survival
probability, and \(\eta(x)\) is the linear predictor (prognostic index).

The linear predictor is
\begin{align}
\eta(x) ={}&
\beta_0
+ \beta_{\text{age}} \,\tilde{\text{age}}
+ \beta_{\text{eth,M\={a}ori}} \, I_{\text{M\={a}ori}}
+ \beta_{\text{eth,Pacific}} \, I_{\text{Pacific}}
\nonumber\\
&{}
+ \beta_{\text{eth,Indian}} \, I_{\text{Indian}}
+ \beta_{\text{eth,Asian}} \, I_{\text{Asian}}
+ \beta_{\text{NZDep}} \,\text{NZDep}
\nonumber\\
&{}
+ \beta_{\text{smk,ex}} \, I_{\text{ex-smoker}}
+ \beta_{\text{smk,current}} \, I_{\text{current-smoker}}
\nonumber\\
&{}
+ \beta_{\text{FamHx}} \, I_{\text{FamHx}}
+ \beta_{\text{AF}} \, I_{\text{AF}}
+ \beta_{\text{DM}} \, I_{\text{diabetes}}
\nonumber\\
&{}
+ \beta_{\text{SBP}} \,\widetilde{\text{SBP}}
+ \beta_{\text{TC/HDL}} \,\widetilde{\text{TC/HDL}}
+ \beta_{\text{BPmed}} \, I_{\text{BP-meds}}
\nonumber\\
&{}
+ \beta_{\text{LIPmed}} \, I_{\text{lipid-meds}}
+ \beta_{\text{ATmed}} \, I_{\text{antithrombotic-meds}}
\nonumber\\
&{}
+ \beta_{\text{age}\times\text{DM}} \,\tilde{\text{age}} \cdot I_{\text{diabetes}}
+ \beta_{\text{age}\times\text{SBP}} \,\tilde{\text{age}} \cdot \widetilde{\text{SBP}}
\nonumber\\
&{}
+ \beta_{\text{BPmed}\times\text{SBP}} \, I_{\text{BP-meds}} \cdot \widetilde{\text{SBP}}.
\end{align}

\paragraph{Predictor definitions.}
The full set of predictors \(X\) is:
\begin{itemize}
  \item Continuous (centred):
    \begin{itemize}
      \item \(\tilde{\text{age}} = \text{age} - \mu_{\text{age}}^{\text{sex}}\)
      \item \(\widetilde{\text{SBP}} = \text{SBP} - \mu_{\text{SBP}}^{\text{sex}}\)
      \item \(\widetilde{\text{TC/HDL}} = \text{TC/HDL} - \mu_{\text{TC/HDL}}^{\text{sex}}\)
    \end{itemize}
  \item Ethnicity indicators (reference: European/Other):
    \begin{itemize}
      \item \(I_{\text{M\={a}ori}} = 1\) if M\={a}ori, 0 otherwise
      \item \(I_{\text{Pacific}} = 1\) if Pacific, 0 otherwise
      \item \(I_{\text{Indian}} = 1\) if Indian, 0 otherwise
      \item \(I_{\text{Asian}} = 1\) if Chinese/other Asian, 0 otherwise
    \end{itemize}
  \item Socioeconomic status:
    \begin{itemize}
      \item \(\text{NZDep} \in \{1,\dots,5\}\): NZ Deprivation Index quintile
    \end{itemize}
  \item Smoking status (reference: never-smoker):
    \begin{itemize}
      \item \(I_{\text{ex-smoker}} = 1\) if ex-smoker, 0 otherwise
      \item \(I_{\text{current-smoker}} = 1\) if current smoker, 0 otherwise
    \end{itemize}
  \item Clinical history:
    \begin{itemize}
      \item \(I_{\text{FamHx}} = 1\) if family history of premature CVD, 0 otherwise
      \item \(I_{\text{AF}} = 1\) if atrial fibrillation, 0 otherwise
      \item \(I_{\text{diabetes}} = 1\) if diabetes, 0 otherwise
    \end{itemize}
  \item Medication indicators:
    \begin{itemize}
      \item \(I_{\text{BP-meds}} = 1\) if on BP-lowering medication, 0 otherwise
      \item \(I_{\text{lipid-meds}} = 1\) if on lipid-lowering medication, 0 otherwise
      \item \(I_{\text{antithrombotic-meds}} = 1\) if on antithrombotic medication, 0 otherwise
    \end{itemize}
  \item Interaction terms:
    \begin{itemize}
      \item \(\tilde{\text{age}} \cdot I_{\text{diabetes}}\) (age \(\times\) diabetes)
      \item \(\tilde{\text{age}} \cdot \widetilde{\text{SBP}}\) (age \(\times\) SBP)
      \item \(I_{\text{BP-meds}} \cdot \widetilde{\text{SBP}}\) (BP meds \(\times\) SBP)
    \end{itemize}
\end{itemize}

\subsection{Applying the Explainability Criterion}

We demonstrate that PREDICT meets the criterion for inherent explainability by constructing a well-formed global explanation with full structural and compositional coverage.

As a reminder, the criterion is as follows:

\explainability*
\subsubsection{General Mechanics}

Two aspects of this model's structure warrant explanation before presenting the annotations: the relationship between coefficients and hazard ratios, and the role of mean-centering.

\paragraph{Coefficients and Hazard Ratios.}
The Cox proportional hazards model expresses the hazard (instantaneous event rate) as:
\[
h(t|\mathbf{x}) = h_0(t) \times \exp(\eta)
\]
where $\eta = \sum \beta_i X_i$ is the linear predictor. Each coefficient $\beta_i$ corresponds to a hazard ratio $\text{HR}_i = \exp(\beta_i)$ for a one-unit change in $X_i$, holding other variables constant. The hazard ratio represents the multiplicative change in the instantaneous event rate:
\begin{itemize}
    \item HR $= 1.10$: the hazard is 10\% higher
    \item HR $= 0.90$: the hazard is 10\% lower
\end{itemize}
For rare events, the hazard ratio approximates the risk ratio, so HR $= 1.10$ corresponds to approximately 10\% higher absolute risk. This approximation holds well for the typical CVD risk ranges in this model (1--20\%). The model is parameterised in $\beta_i$ coefficients, but we present interpretations in terms of hazard ratios as these are more clinically intuitive.

\paragraph{Mean-Centering.}
Continuous variables (age, SBP, TC/HDL) are centered at their sex-specific population means: $\tilde{X}_i = X_i - \mu_i^{\text{sex}}$. This transformation does not change the model's predictions but aids interpretability. For an individual at the population mean on all continuous variables, $\tilde{X}_i = 0$, so their contribution to the linear predictor from these variables is zero. Consequently, $\exp(\eta) = 1$ and the survival simplifies to $S(5) = S_0(5)^1 = S_0(5)$. This makes $S_0(5)$ directly interpretable as the survival probability for a person of average age, blood pressure, and cholesterol (at reference categories for all other variables). Without centering, $S_0(5)$ would correspond to someone with age $= 0$ and blood pressure $= 0$---a meaningless reference point.
\subsubsection{Structural Coverage: Leaf Annotations}

We create 12 leaf annotations, one for each predictor variable. Each annotation follows the hypothesis-evidence structure (Definition~\ref{def:hypothesis-evidence}).

\paragraph{Annotation $A_1$: Age.}
\begin{itemize}
    \item \textbf{Subgraph:} Input node ``age'' $\rightarrow$ centering transformation $\rightarrow$ coefficient $\beta_{\text{age}}$ $\rightarrow$ linear predictor
    \item \textbf{Hypothesis:} ``Each additional year of age increases 5-year CVD risk by approximately 7--8\% in relative terms (hazard ratio $\sim$1.07--1.08 per year). This effect is modified by diabetes status and blood pressure.''
    \item \textbf{Evidence (Analytical):} The contribution to the linear predictor is:
    \[
    \eta_{\text{age}} = \beta_{\text{age}} \cdot (\text{age} - \mu_{\text{age}}^{\text{sex}})
    \]
    \begin{itemize}
        \item Women: $\beta_{\text{age}} = 0.0756$, HR $= \exp(0.0756) = 1.078$ per year
        \item Men: $\beta_{\text{age}} = 0.0676$, HR $= \exp(0.0676) = 1.070$ per year
    \end{itemize}
    \item \textbf{Notes:} Age alone achieves C-statistic $= 0.69$. Removing age reduces model $R^2$ from 0.30 to 0.08.
\end{itemize}

\paragraph{Annotation $A_2$: Ethnicity.}
\begin{itemize}
    \item \textbf{Subgraph:} One-hot encoded ethnicity indicators $\rightarrow$ coefficients ($\beta_{\text{eth},k}$) $\rightarrow$ linear predictor
    \item \textbf{Hypothesis:} ``Compared to Europeans, M\=aori and Indian patients have 13--48\% higher 5-year CVD risk, Pacific peoples have 19--22\% higher risk, while Chinese and other Asian populations have 25--33\% lower risk, after adjusting for all other risk factors.''
    \item \textbf{Evidence (Analytical):} The contribution to the linear predictor is:
    \[
    \eta_{\text{eth}} = \beta_{\text{M\={a}ori}} I_{\text{M\={a}ori}} + \beta_{\text{Pacific}} I_{\text{Pacific}} + \beta_{\text{Indian}} I_{\text{Indian}} + \beta_{\text{Asian}} I_{\text{Asian}}
    \]
    Hazard ratios vs.\ European reference (women / men):
    \begin{itemize}
        \item M\=aori: HR $= 1.48$ / $1.34$
        \item Pacific: HR $= 1.22$ / $1.19$
        \item Indian: HR $= 1.13$ / $1.34$
        \item Chinese/Asian: HR $= 0.75$ / $0.67$
    \end{itemize}
    \item \textbf{Notes:} Ethnicity effects remain significant after adjustment for deprivation, indicating pathways beyond socioeconomic factors.
\end{itemize}

\paragraph{Annotation $A_3$: Socioeconomic Deprivation (NZDep).}
\begin{itemize}
    \item \textbf{Subgraph:} Input node ``NZDep quintile'' $\rightarrow$ coefficient $\beta_{\text{NZDep}}$ $\rightarrow$ linear predictor
    \item \textbf{Hypothesis:} ``Each quintile increase in socioeconomic deprivation (from least to most deprived) increases 5-year CVD risk by approximately 9--10\% in relative terms. This gradient operates independently of clinical risk factors.''
    \item \textbf{Evidence (Analytical):} The contribution to the linear predictor is:
    \[
    \eta_{\text{dep}} = \beta_{\text{NZDep}} \cdot (\text{NZDep} - 3)
    \]
    where NZDep is centered at quintile 3.
    \begin{itemize}
        \item Women: $\beta_{\text{NZDep}} = 0.108$, HR $= 1.11$ per quintile (95\% CI 1.09--1.14)
        \item Men: $\beta_{\text{NZDep}} = 0.079$, HR $= 1.08$ per quintile (95\% CI 1.07--1.10)
    \end{itemize}
    \item \textbf{Notes:} NZDep is an area-based measure incorporating income, employment, education, housing, and access to services. The gradient implies that moving from quintile 1 (least deprived) to quintile 5 (most deprived) increases risk by approximately 40\%.
\end{itemize}

\paragraph{Annotation $A_4$: Smoking Status.}
\begin{itemize}
    \item \textbf{Subgraph:} Categorical input ``smoking status'' (never/ex/current) $\rightarrow$ indicator coefficients $\rightarrow$ linear predictor
    \item \textbf{Hypothesis:} ``Current smoking increases 5-year CVD risk by 66--86\% compared to never smokers. Ex-smokers retain 15--25\% elevated risk, reflecting partial but incomplete reversal of smoking-related damage.''
    \item \textbf{Evidence (Analytical):} The contribution to the linear predictor is:
    \[
    \eta_{\text{smk}} = \beta_{\text{ex}} \cdot I_{\text{ex-smoker}} + \beta_{\text{current}} \cdot I_{\text{current}}
    \]
    Hazard ratios vs.\ never-smoker reference (women / men):
    \begin{itemize}
        \item Ex-smoker: HR $= 1.09$ / $1.08$   
        \item Current smoker: HR $= 1.86$ / $1.66$
    \end{itemize}
    \item \textbf{Notes:} Smoking effects are not modified by age or other variables in this model. The ex-smoker category does not distinguish by duration of cessation.
\end{itemize}

\paragraph{Annotation $A_5$: Systolic Blood Pressure (SBP).}
\begin{itemize}
    \item \textbf{Subgraph:} Input node ``SBP'' $\rightarrow$ centering transformation $\rightarrow$ main effect coefficient + interaction terms $\rightarrow$ linear predictor
    \item \textbf{Hypothesis:} ``Each 10 mmHg increase in systolic blood pressure increases 5-year CVD risk by 15--18\% in relative terms. This effect is attenuated by increasing age and by use of BP-lowering medications.''
    \item \textbf{Evidence (Analytical):} The contribution to the linear predictor is:
    \[
    \eta_{\text{SBP}} = \beta_{\text{SBP}} \cdot \widetilde{\text{SBP}} + \beta_{\text{age} \times \text{SBP}} \cdot \tilde{\text{age}} \cdot \widetilde{\text{SBP}} + \beta_{\text{BPmed} \times \text{SBP}} \cdot I_{\text{BP-meds}} \cdot \widetilde{\text{SBP}}
    \]
    where $\widetilde{\text{SBP}} = \text{SBP} - \mu_{\text{SBP}}^{\text{sex}}$ (mean $\approx 129$ mmHg).
    
    Main effect (women / men):
    \begin{itemize}
        \item $\beta_{\text{SBP}} = 0.0137$ / $0.0164$
        \item HR per 10 mmHg $= 1.15$ / $1.18$
    \end{itemize}
    
    Interaction effects (women / men):
    \begin{itemize}
        \item Age $\times$ SBP: $\beta = -0.00044$ / $-0.00042$ (attenuates SBP effect at older ages)
        \item BP meds $\times$ SBP: $\beta = -0.0043$ / $-0.0053$ (attenuates SBP effect in treated patients)
    \end{itemize}
    \item \textbf{Notes:} The age interaction reflects declining relative (but not absolute) risk from hypertension with age. The medication interaction captures effect modification: BP medications reduce marginal risk per mmHg.
\end{itemize}

\paragraph{Annotation $A_6$: Total Cholesterol/HDL Ratio.}
\begin{itemize}
    \item \textbf{Subgraph:} Input node ``TC/HDL ratio'' $\rightarrow$ centering transformation $\rightarrow$ coefficient $\beta_{\text{TC/HDL}}$ $\rightarrow$ linear predictor
    \item \textbf{Hypothesis:} ``Each 1-unit increase in the TC/HDL ratio increases 5-year CVD risk by approximately 13--14\% in relative terms. This effect operates independently without interaction terms.''
    \item \textbf{Evidence (Analytical):} The contribution to the linear predictor is:
    \[
    \eta_{\text{TC/HDL}} = \beta_{\text{TC/HDL}} \cdot (\text{TC/HDL} - \mu_{\text{TC/HDL}}^{\text{sex}})
    \]
    where mean TC/HDL $\approx 3.7$ (women) / $4.4$ (men).
    \begin{itemize}
        \item Women: $\beta_{\text{TC/HDL}} = 0.122$, HR $= \exp(0.122) = 1.13$ per unit
        \item Men: $\beta_{\text{TC/HDL}} = 0.131$, HR $= \exp(0.131) = 1.14$ per unit
    \end{itemize}
    \item \textbf{Notes:} Fractional polynomials confirmed linear relationship with log-hazard. No significant interaction with lipid-lowering medication in this model.
\end{itemize}

\paragraph{Annotation $A_7$: Diabetes Status.}
\begin{itemize}
    \item \textbf{Subgraph:} Input node ``diabetes'' $\rightarrow$ main effect coefficient + age interaction $\rightarrow$ linear predictor
    \item \textbf{Hypothesis:} ``Diabetes increases CVD risk by 72--75\% at mean age, but this relative effect diminishes with increasing age. At age 40, diabetes confers HR $\sim$2.0; at age 70, diabetes confers HR $\sim$1.4.''
    \item \textbf{Evidence (Analytical):} The contribution to the linear predictor is:
    \[
    \eta_{\text{DM}} = \beta_{\text{DM}} \cdot I_{\text{diabetes}} + \beta_{\text{age} \times \text{DM}} \cdot \tilde{\text{age}} \cdot I_{\text{diabetes}}
    \]
    Coefficients (women / men):
    \begin{itemize}
        \item Main effect $\beta_{\text{DM}}$: $0.544$ / $0.560$, HR at mean age $= 1.72$ / $1.75$
        \item Age interaction $\beta_{\text{age} \times \text{DM}}$: $-0.0223$ / $-0.0202$
    \end{itemize}
    Combined HR by age (evaluated at mean of other covariates):
    \begin{itemize}
        \item Age 40: HR$_{\text{diabetes}} \approx 2.04$ (women) / $1.98$ (men)
        \item Age 55 (mean): HR$_{\text{diabetes}} \approx 1.72$ / $1.75$
        \item Age 70: HR$_{\text{diabetes}} \approx 1.43$ / $1.48$
    \end{itemize}
    \item \textbf{Notes:} The declining relative risk with age reflects that baseline CVD risk increases with age, so the proportional contribution of diabetes decreases even as absolute risk contribution may remain constant.
\end{itemize}

\paragraph{Annotation $A_8$: Atrial Fibrillation.}
\begin{itemize}
    \item \textbf{Subgraph:} Input node ``atrial fibrillation diagnosis'' $\rightarrow$ coefficient $\beta_{\text{AF}}$ $\rightarrow$ linear predictor
    \item \textbf{Hypothesis:} ``A history of atrial fibrillation increases 5-year CVD risk by 80--144\% in relative terms. This is a strong independent risk factor, particularly for stroke, with no age interaction in the published model.''
    \item \textbf{Evidence (Analytical):} The contribution to the linear predictor is:
    \[
    \eta_{\text{AF}} = \beta_{\text{AF}} \cdot I_{\text{AF}}
    \]
    Coefficients (women / men):
    \begin{itemize}
        \item Women: $\beta_{\text{AF}} = 0.893$, HR $= \exp(0.893) = 2.44$
        \item Men: $\beta_{\text{AF}} = 0.588$, HR $= \exp(0.588) = 1.80$
    \end{itemize}
    \item \textbf{Notes:} AF is a particularly strong risk factor for stroke. The prevalence is low ($\sim$1--2\% of cohort) but the hazard ratio is substantial. Unlike diabetes, there is no age interaction term for AF in the published PREDICT model.
\end{itemize}

\paragraph{Annotation $A_9$: Family History of Premature CVD.}
\begin{itemize}
    \item \textbf{Subgraph:} Input node ``family history'' (binary) $\rightarrow$ coefficient $\beta_{\text{FamHx}}$ $\rightarrow$ linear predictor
    \item \textbf{Hypothesis:} ``A family history of premature CVD (first-degree relative with CVD before age 55 in men or 65 in women) increases 5-year CVD risk by approximately 5--14\%, with the effect being statistically significant only in men.''
    \item \textbf{Evidence (Analytical):} The contribution to the linear predictor is:
    \[
    \eta_{\text{FamHx}} = \beta_{\text{FamHx}} \cdot I_{\text{FamHx}}
    \]
    \begin{itemize}
        \item Women: $\beta_{\text{FamHx}} = 0.045$, HR $= \exp(0.045) = 1.05$ (not statistically significant)
        \item Men: $\beta_{\text{FamHx}} = 0.133$, HR $= \exp(0.133) = 1.14$
    \end{itemize}
    \item \textbf{Notes:} Family history captures genetic predisposition and shared environmental/behavioral factors not otherwise measured. No interaction terms in model.
\end{itemize}

\paragraph{Annotation $A_{10}$: Blood Pressure-Lowering Medications.}
\begin{itemize}
    \item \textbf{Subgraph:} Input node ``BP medication use'' $\rightarrow$ main effect coefficient + interaction terms $\rightarrow$ linear predictor
    \item \textbf{Hypothesis:} ``Use of BP-lowering medications is associated with 30--40\% higher CVD risk, reflecting \textbf{confounding by indication}---patients are prescribed these medications because they already have elevated risk. The interaction with SBP shows that medication \textbf{modifies the SBP-risk relationship}, reducing the marginal risk per mmHg.''
    \item \textbf{Evidence (Analytical):} The contribution to the linear predictor is:
    \[  
    \eta_{\text{BPmed}} = \beta_{\text{BPmed}} \cdot I_{\text{BP-meds}} + \beta_{\text{BPmed} \times \text{SBP}} \cdot I_{\text{BP-meds}} \cdot \widetilde{\text{SBP}}
    \]
    
    Coefficients (women / men):
    \begin{itemize}
        \item Main effect $\beta_{\text{BPmed}}$: $0.340$ / $0.295$, HR $= 1.40$ / $1.34$
        \item SBP interaction: $\beta = -0.0043$ / $-0.0053$ (treatment modifies SBP slope)
    \end{itemize}
    \item \textbf{Notes:} The positive main effect does not mean medications increase risk---it reflects that treated patients have higher underlying risk. The negative SBP interaction is the clinically relevant finding: at any given SBP, treated patients have lower marginal risk per mmHg than untreated patients.
\end{itemize}

\paragraph{Annotation $A_{11}$: Lipid-Lowering Medications.}
\begin{itemize}
    \item \textbf{Subgraph:} Input node ``lipid medication use'' $\rightarrow$ coefficient $\beta_{\text{LIPmed}}$ $\rightarrow$ linear predictor
    \item \textbf{Hypothesis:} ``Use of lipid-lowering medications is not significantly associated with CVD risk in this model (HR $\approx$ 0.94--0.95, confidence intervals crossing 1.0). This null finding may reflect effective treatment offsetting the confounding by indication seen with other medication classes, or insufficient statistical power.''
    \item \textbf{Evidence (Analytical):} The contribution to the linear predictor is:
    \[
    \eta_{\text{LIPmed}} = \beta_{\text{LIPmed}} \cdot I_{\text{lipid-meds}}
    \]
    \begin{itemize}
        \item Women: $\beta_{\text{LIPmed}} = -0.059$, HR $= \exp(-0.059) = 0.94$ (95\% CI 0.88--1.01)
        \item Men: $\beta_{\text{LIPmed}} = -0.054$, HR $= \exp(-0.054) = 0.95$ (95\% CI 0.90--1.00)
    \end{itemize}
    \item \textbf{Notes:} Unlike BP and antithrombotic medications, lipid-lowering medication use shows no significant positive association with CVD risk. There is no significant lipid-medication $\times$ TC/HDL interaction in the PREDICT model.
\end{itemize}
\paragraph{Annotation $A_{12}$: Antithrombotic Medications.}
\begin{itemize}
    \item \textbf{Subgraph:} Input node ``antithrombotic use'' $\rightarrow$ coefficient $\beta_{\text{ATmed}}$ $\rightarrow$ linear predictor
    \item \textbf{Hypothesis:} ``Use of antiplatelet or anticoagulant medications is associated with 10--12\% higher CVD risk, reflecting confounding by indication. Patients prescribed these agents typically have prior indicators of vascular risk (e.g., transient ischemic symptoms, peripheral vascular disease not meeting exclusion criteria).''

    \item \textbf{Evidence (Analytical):} The contribution to the linear predictor is:
    \[
    \eta_{\text{ATmed}} = \beta_{\text{ATmed}} \cdot I_{\text{antithrombotic}}
    \]
    \begin{itemize}
        \item Women: $\beta_{\text{ATmed}} = 0.117$, HR $= \exp(0.117) = 1.12$
        \item Men: $\beta_{\text{ATmed}} = 0.093$, HR $= \exp(0.093) = 1.10$
    \end{itemize}
    \item \textbf{Notes:} This category includes both antiplatelet agents (aspirin, clopidogrel) and anticoagulants (warfarin). The elevated HR reflects selection of high-risk patients for prophylactic therapy.
\end{itemize}

\paragraph{Structural Coverage Assessment:} All 12 predictor variables have been annotated with hypothesis-evidence structures. Every input node in the computational graph is covered by exactly one leaf annotation: $C_V^{\text{struct}} = 1.0$.

\subsubsection{Compositional Coverage}

We construct composition annotations explaining how the 12 leaf annotations combine into a global explanation through clinically meaningful groupings.

\paragraph{Level 1: Clinical Groupings.}

\textbf{Composition $C_1$: Demographics Cluster.}
\begin{itemize}
    \item \textbf{Components:} $A_1$ (Age), $A_2$ (Ethnicity), $A_3$ (Deprivation), $A_9$ (Family History)
    \item \textbf{Hypothesis:} ``Demographic factors---age, ethnicity, socioeconomic status, and family history---combine \textbf{additively on the log-hazard scale}. These factors establish baseline risk levels before considering modifiable clinical factors. Age contributes the largest effect; ethnicity and deprivation capture social determinants of health; family history captures unmeasured genetic and shared environmental factors.''
    \item \textbf{Evidence (Analytical):}
    \[
    \eta_{\text{demographics}} = \eta_{\text{age}} + \eta_{\text{ethnicity}} + \eta_{\text{deprivation}} + \eta_{\text{FamHx}}
    \]
    The Cox model structure specifies additive combination on the log-hazard scale. All coefficients remain significant ($p<0.001$) in the multivariable model, confirming independent contributions. No interaction terms exist among demographic variables.
\end{itemize}

\textbf{Composition $C_2$: Clinical Biomarkers Cluster.}
\begin{itemize}
    \item \textbf{Components:} $A_5$ (SBP), $A_6$ (TC/HDL)
    \item \textbf{Hypothesis:} ``Systolic blood pressure and lipid ratio combine \textbf{additively on the log-hazard scale} to represent cardiovascular physiological burden. These two biomarkers capture distinct pathophysiological pathways---hemodynamic stress (SBP) and atherogenic lipid profile (TC/HDL)---that contribute independently to CVD risk.''
    \item \textbf{Evidence (Analytical):}
    \[
    \eta_{\text{biomarkers}} = \eta_{\text{SBP}} + \eta_{\text{TC/HDL}}
    \]
    Expanding:
    \begin{align*}
    \eta_{\text{biomarkers}} &= \beta_{\text{SBP}} \cdot \widetilde{\text{SBP}} + \beta_{\text{TC/HDL}} \cdot \widetilde{\text{TC/HDL}} \\
    &\quad + \beta_{\text{age} \times \text{SBP}} \cdot \tilde{\text{age}} \cdot \widetilde{\text{SBP}} \\
    &\quad + \beta_{\text{BPmed} \times \text{SBP}} \cdot I_{\text{BP-meds}} \cdot \widetilde{\text{SBP}}
    \end{align*}
    Verification:
    \begin{itemize}
        \item Additive structure confirmed by model specification
        \item Fractional polynomials analysis found no significant non-linear interaction between SBP and TC/HDL
        \item No SBP $\times$ TC/HDL interaction term in the model (tested, $p > 0.05$)
        \item SBP interactions with age and medication are \emph{within} the SBP component, not cross-biomarker
    \end{itemize}
\end{itemize}

\textbf{Composition $C_3$: Behavioral and Comorbidity Cluster.}
\begin{itemize}
    \item \textbf{Components:} $A_4$ (Smoking), $A_7$ (Diabetes), $A_8$ (Atrial Fibrillation)
    \item \textbf{Hypothesis:} ``Smoking, diabetes, and atrial fibrillation combine \textbf{additively on the log-hazard scale}, representing distinct but cumulative pathways to CVD. Smoking acts through endothelial dysfunction and thrombosis; diabetes through metabolic and microvascular damage; AF through cardioembolism and hemodynamic effects. The diabetes effects are \textbf{modified by age} (interaction terms within each component).''
    \item \textbf{Evidence (Analytical):}
    \[
    \eta_{\text{behavioral}} = \eta_{\text{smoking}} + \eta_{\text{diabetes}} + \eta_{\text{AF}}
    \]
    Composition structure:
    \begin{itemize}
        \item Smoking: purely additive main effect (no interactions)
        \item Diabetes: additive with age-modification via $\beta_{\text{age} \times \text{DM}}$
        \item AF: purely additive main effect (no interactions)
        \item No cross-component interactions (e.g., no smoking $\times$ diabetes term)
    \end{itemize}
\end{itemize}

\textbf{Composition $C_4$: Treatment Modification Cluster.}
\begin{itemize}
    \item \textbf{Components:} $A_{10}$ (BP Medications), $A_{11}$ (Lipid Medications), $A_{12}$ (Antithrombotic Medications)
    \item \textbf{Hypothesis:} ``Medication use modifies risk through two mechanisms: (1) \textbf{confounding by indication}---patients on treatment have higher baseline risk because they were prescribed medications for elevated risk factors; (2) \textbf{effect modification}---BP medications reduce the marginal risk per mmHg of SBP. Medications combine additively, with the BP medication $\times$ SBP interaction representing the effect modification mechanism.''
    \item \textbf{Evidence (Analytical):}
    \[
    \eta_{\text{treatment}} = \eta_{\text{BPmed}} + \eta_{\text{LIPmed}} + \eta_{\text{ATmed}}
    \]
    Key interaction:
    \begin{itemize}
        \item For untreated patients: SBP effect $= \beta_{\text{SBP}}$
        \item For BP-treated patients: SBP effect $= \beta_{\text{SBP}} + \beta_{\text{BPmed} \times \text{SBP}}$
        \item The negative $\beta_{\text{BPmed} \times \text{SBP}}$ reduces marginal risk per mmHg in treated patients
    \end{itemize}
\end{itemize}

\paragraph{Level 2: Global Composition.}

\textbf{Composition $C_{\text{global}}$: Complete Model.}
\begin{itemize}
    \item \textbf{Components:} $C_1$ (Demographics), $C_2$ (Biomarkers), $C_3$ (Behavioral/Comorbidities), $C_4$ (Treatment)
    \item \textbf{Hypothesis:} ``All four clinical clusters combine \textbf{additively on the log-hazard scale} to form the complete prognostic index. Absolute 5-year risk is then calculated via the Cox survival function transformation.''
    \item \textbf{Evidence (Analytical):}
    \begin{equation}
    \begin{split}
        \log h(t|\mathbf{x}) = \log h_0(t) &+ \underbrace{\eta_{\text{demographics}}}_{C_1} + \underbrace{\eta_{\text{biomarkers}}}_{C_2} \\
        &+ \underbrace{\eta_{\text{behavioral}}}_{C_3} + \underbrace{\eta_{\text{treatment}}}_{C_4}
    \end{split}
    \end{equation}
    Absolute risk transformation:
    \[
    \text{Risk}_{5\text{yr}} = 1 - S_0(5)^{\exp(\eta)}
    \]
    where baseline survival $S_0(5) = 0.9832$ (women) / $0.9748$ (men).
    
    Model performance:
    \begin{itemize}
        \item Calibration: predicted vs.\ observed slope $= 0.98$
        \item Discrimination: C-statistic $= 0.73$ for both sexes
        \item Internal validation: bootstrap-corrected optimism $< 0.01$
    \end{itemize}
\end{itemize}

\paragraph{Note on Cross-Cluster Interactions.}

The PREDICT model includes three interaction terms, two of which span cluster boundaries:

\begin{enumerate}
    \item \textbf{Age $\times$ Diabetes} ($C_1 \times C_3$): The demographic factor ``age'' modifies the comorbidity factor ``diabetes.'' This interaction is captured \emph{within} the diabetes annotation ($A_7$) rather than as a separate cross-cluster term.
    \item \textbf{Age $\times$ SBP} ($C_1 \times C_2$): Age modifies the biomarker SBP. This is captured \emph{within} the SBP annotation ($A_5$).
    \item \textbf{BP Medication $\times$ SBP} ($C_4 \times C_2$): Treatment modifies biomarker effect. This is captured in \emph{both} $A_5$ (as part of the SBP contribution) and $A_{10}$ (as the mechanism of effect modification).
\end{enumerate}

These cross-cluster interactions do not violate the additive composition structure because they are \textbf{modeled as part of the individual predictor contributions}. Mathematically, the interaction $\beta_{\text{age} \times \text{DM}} \cdot \tilde{\text{age}} \cdot I_{\text{diabetes}}$ is assigned entirely to $\eta_{\text{diabetes}}$ rather than split between clusters. This assignment is both mathematically valid (the total linear predictor is unchanged) and semantically appropriate (the interaction describes how diabetes risk varies with age, so it belongs with the diabetes explanation).

\paragraph{Compositional Coverage Assessment:} We have constructed a 3-level annotation hierarchy: 12 leaf annotations (individual predictors), 4 composition annotations (clinical clusters), and 1 global composition annotation (complete model). $C_V^{\text{comp}} = 1.0$.
\subsubsection{Well-Formed Global Explanation}

We have demonstrated:
\begin{enumerate}
    \item All leaf annotations are valid (each has hypothesis, evidence, and verification)
    \item Full structural coverage achieved ($C_V^{\text{struct}} = 1.0$)
    \item Full compositional coverage achieved ($C_V^{\text{comp}} = 1.0$)
    \item Root annotation covers global model (entry = 12 inputs, exit = 5-year risk)
\end{enumerate}

\textbf{Conclusion:} The PREDICT CVD risk model meets the explainability criterion (Criterion 1) and is \textbf{inherently explainable}.

\subsection{Insights from the Case Study}

\paragraph{Time and Expertise Requirements.} The base of this explanation was produced by Perplexity with about 2 minutes of prompting and 15 minutes of thinking time. Approximately two hours were spent refining the initial explanation into the form as presented. Approximately [INSERT] hours were spent doing a close review of the explanations, checking all claims for accuracy, and verifying each annotation against the original papers of the PREDICT team. The authors have prior familiarity with the PREDICT model which has allowed this to be done relatively efficiently.

\paragraph{Tractability.} The use of tools such as GenAI we consider to be entirely valid in the production of explanations. The criterion has no requirements on the methods to produce an explanation, only a standard by which an explanation can be validated. We consider it to be a strength that these new AI tools can verifiably create explanations that can help improve transparency in a way that can support regulatory governance. The basis for trust here is in the verification process, and in the same way that GenAI is not considered an epistemically rigorous form of peer review, similarly we do not suggest that GenAI be used as the basis for verification of explanations, even if they are used in the creation of them. GenAI is not the audience that matters - we are.

\paragraph{Regulatory Implications.}

The PREDICT model is officially endorsed by the New Zealand Ministry of Health and implemented in the national CVD risk assessment standard (HISO 10071:2019). This case study demonstrates how the explainability criterion could be used in regulatory frameworks:
\begin{enumerate}
    \item \textbf{Requirement:} Model creators must provide well-formed explanation
    \item \textbf{Submission:} Annotation hierarchy with hypothesis-evidence structures
    \item \textbf{Review:} Regulators verify that evidence meets epistemic standards
    \item \textbf{Outcome:} If criterion met, model approved; if not, rejected or limited to research
\end{enumerate}

\subsection{Case Study Conclusion}

The PREDICT cardiovascular disease risk model \textbf{demonstrably meets the explainability criterion} through full structural coverage (12 leaf annotations), tractable compositional coverage (3-level hierarchy with clinically meaningful groupings), verified hypothesis-evidence structures for all annotations, and multi-audience explanations. This validates the framework on a real-world clinical decision support system serving hundreds of thousands of patients, demonstrating practical applicability beyond toy examples.

\section{Worked Examples}

\subsection{Linear Regression}

\textbf{Model}: $y = 2x_1 + 3x_2 + 1$

\textbf{Graph}: 
\begin{itemize}
    \item $V = \{x_1, x_2, y\}$
    \item $E = \{(x_1, y), (x_2, y)\}$
\end{itemize}

\textbf{Annotations}:
\begin{itemize}
    \item $A_1$: Entry $\{x_1\}$, Exit $\{y\}$, $E_{A_1} = \{(x_1, y)\}$
    \item $A_2$: Entry $\{x_2\}$, Exit $\{y\}$, $E_{A_2} = \{(x_2, y)\}$
\end{itemize}

\textbf{Coverage}: All nodes covered. $C_V^{\text{struct}} = 1.0$ \checkmark

\textbf{Composition}: ``Features combine additively.'' $C_V^{\text{comp}} = 1.0$ \checkmark

\subsection{Branching Structure (Partial Coverage)}

\textbf{Graph}:
\begin{itemize}
    \item $V = \{I, H_1, H_2, O\}$
    \item $E = \{(I, H_1), (I, H_2), (H_1, O), (H_2, O)\}$
\end{itemize}

\textbf{Annotation} (covering only the $H_1$ path):
\begin{itemize}
    \item $A_1$: $V_{A_1} = \{I, H_1, O\}$, $E_{A_1} = \{(I, H_1), (H_1, O)\}$
\end{itemize}

\textbf{Coverage}:
\begin{itemize}
    \item $I$: $E_{\text{out}}(I) = \{(I, H_1), (I, H_2)\}$, only $(I, H_1) \in E_{A_1}$ $\rightarrow$ \textbf{not covered}
    \item $H_1$: covered \checkmark
    \item $H_2$: not in $V_{A_1}$ $\rightarrow$ \textbf{not covered}
    \item $O$: covered \checkmark
\end{itemize}

$C_V^{\text{struct}} = 2/4 = 0.50$

\textbf{To achieve full coverage}: Add annotation $A_2$ covering the $H_2$ path.

\subsection{Node Splitting Example}

\textbf{Graph with dual-function node:}
\begin{verbatim}
    I1 ---> H ---> O1
            \
             -> O2
\end{verbatim}

Node $H$ sends output to both $O_1$ and $O_2$. An annotation covering only $\{I_1, H, O_1\}$ would be invalid because $E_{\text{out}}(H) = \{(H, O_1), (H, O_2)\} \not\subseteq E_A$.

\textbf{After splitting $H$ into $H_a$ and $H_b$:}
\begin{verbatim}
    I1 ---> Ha ---> O1
      \
       -> Hb ---> O2
\end{verbatim}

Now annotation $A_1 = \{I_1, H_a, O_1\}$ with $E_{A_1} = \{(I_1, H_a), (H_a, O_1)\}$ is valid.

\subsubsection{Node Splitting for Dual-Function Nodes}

When a node serves multiple functions - sending outputs both within a subgraph and outside it - we can resolve this through \textbf{node splitting}. The single node is conceptually replaced by two nodes that share the same incoming computation but have distinct outgoing connections.

\paragraph{Example:} If node $H_1$ sends output both to $O$ (within the subgraph) and to $X$ (outside), we split it:
\begin{itemize}
    \item $H_{1a}$: receives same inputs, sends output to $O$
    \item $H_{1b}$: receives same inputs, sends output to $X$
\end{itemize}

Now an annotation can validly cover $\{I_1, H_{1a}, O\}$ with $H_{1a}$ as an interior node whose only outgoing edge is to $O$. The node $H_{1b}$ remains outside this annotation and must be covered by a different annotation.

This splitting is conceptual - it represents how we reason about the model for explanation purposes. The underlying model computation is unchanged; we are decomposing our \emph{explanation} of that computation.

\paragraph{Note on entry/exit terminology:} We use ``entry'' and ``exit'' rather than ``input'' and ``output'' because subgraphs may begin and end at intermediate nodes, not just model inputs and outputs.

\end{document}